\documentclass{article}

\PassOptionsToPackage{numbers, compress}{natbib}
  \usepackage[preprint]{neurips_2026}

\usepackage[utf8]{inputenc} 
\usepackage[T1]{fontenc}    
\usepackage{hyperref}       
\usepackage{url}            
\usepackage{booktabs}       
\usepackage{amsmath,amssymb}
\usepackage{nicefrac}       
\usepackage{microtype}      
\usepackage{xcolor}         
\usepackage{bbding}
\definecolor{clr_barrier}{HTML}{FF8C00}
\definecolor{clr_bicycle}{HTML}{FFB6C1}
\definecolor{clr_bus}{HTML}{FFFF00}
\definecolor{clr_car}{HTML}{0099FF}
\definecolor{clr_cveh}{HTML}{00FFFF}
\definecolor{clr_motor}{HTML}{B89400}
\definecolor{clr_ped}{HTML}{FF0000}
\definecolor{clr_tcone}{HTML}{FFF0A0}
\definecolor{clr_trailer}{HTML}{8B4513}
\definecolor{clr_truck}{HTML}{9932CC}
\definecolor{clr_drvsurf}{HTML}{FF00FF}
\definecolor{clr_othflat}{HTML}{808080}
\definecolor{clr_sidewlk}{HTML}{330033}
\definecolor{clr_terrain}{HTML}{90EE90}
\definecolor{clr_manmade}{HTML}{E6E6FA}
\definecolor{clr_veg}{HTML}{00A000}

\usepackage{graphicx}
\usepackage{placeins}
\usepackage{wrapfig}
\usepackage{caption}

\title{VGGT-Occ: Geometry-Grounded and Density-Aware Gated Fusion for 3D Occupancy Prediction}

\author{%
 Xun Chen$^*$\textsuperscript{1}, Tianchen Deng$^*$\Envelope\textsuperscript{2}, Rui Wang\textsuperscript{2}, Fangjinhua Wang\textsuperscript{3}, Junyi Ma\textsuperscript{2}, Hongming Shen\Envelope\textsuperscript{1},\\  \textbf{Hesheng Wang\textsuperscript{2}}, \textbf{Danwei Wang\textsuperscript{1}} \\
{\textsuperscript{\rm 1} Nanyang Technological University}
{\textsuperscript{\rm 2} Shanghai Jiao Tong University}
{\textsuperscript{\rm 3} ETH Zurich}
}

\begin{document}

\maketitle

\begin{abstract}

3D semantic occupancy prediction requires accurate 2D-to-3D feature lifting, yet current methods restrict camera geometry to initial projections. Subsequent operations like offset learning, attention weighting, and cross-camera aggregation remain geometry-agnostic, ignoring essential physical constraints. We propose VGGT-Occ, a framework that embeds geometric tokens throughout the entire pipeline. We introduce Projection-Aware Deformable Attention (PA-DA) to inject geometry into all attention stages. PA-DA projects 3D offsets back to image planes and leverages the projection Jacobian as an additive bias to suppress unreliable observations. Features are then integrated through a view-quality semantic gate for cross-view consistency. To optimize both efficiency and performance, we employ a sequential coarse-to-fine decoder with gated fusion, where low-resolution features are refined into higher resolutions, allocating computation by information density while substantially reducing decoder cost. Extensive evaluations demonstrate the effectiveness and accuracy of our approach. On SurroundOcc-nuScenes, VGGT-Occ achieves 33.00\% IoU and 21.08\% mIoU ($T{=}1$), and 33.64\% IoU and 21.43\% mIoU with $T{=}2$ inference, outperforming existing methods, with only ${\sim}41$M trainable parameters in the occupancy head. Code will be released publicly.

\end{abstract}

\section{Introduction}
\label{sec:intro}

3D semantic occupancy prediction aims to reconstruct the surrounding environment as a dense voxel grid with fine-grained semantic labels. As a cornerstone of autonomous driving perception, it provides a comprehensive geometric and semantic understanding of complex scenes~\cite{zhang2026survey,xu2025survey,surroundocc, occ3d}.
The central challenge is recovering complete 3D structure from multi-camera 2D images, which requires accurate cross-view reasoning, occlusion handling, and precise 2D-to-3D feature lifting.

Despite diverse representation choices~\cite{deng2025best3dscenerepresentation} (dense voxels~\cite{surroundocc, occformer}, sparse Gaussians~\cite{gaussianformer2,deng2024compact,qian2026splatssc, qian2025tgsformer}, or superquadrics~\cite{quadricformer}), prevailing query-based multi-view perception methods suffer from a common architectural flaw: camera geometry is treated as an isolated preprocessing step. Intrinsics and extrinsics are typically restricted to projecting 3D reference points onto 2D feature maps.
Beyond this initial projection, the subsequent deformable cross-attention loop operates in a geometry-blind feature space across three critical stages:
\textbf{(1)~2D Offsets Learning}: Sampling offsets are typically learned in the 2D image plane. Consequently, a unified 3D semantic direction (e.g., "above the vehicle") translates into inconsistent 2D pixel shifts across different views, breaking cross-view geometric consistency.
Sparse4D~\cite{sparse4d} and Far3D~\cite{jiang2024far3d} address this via 3D offset learning with projection, but do not extend projection geometry to attention weighting or cross-camera aggregation (stages~(2) and~(3)).
\textbf{(2)~Blind Attention Weighting}: 
In standard cross-view deformable attention, attention weights are generated purely from 3D queries. The network is fundamentally blind to the perspective distortion and resolution degradation that occur during 3D-to-2D projection. A query might blindly assign high weight to a sampled point that suffers from severe perspective distortion (e.g., far depth or large incidence angle).
\textbf{(3)~Naive Cross-Camera Averaging}: Features sampled from different cameras are simply averaged, disregarding the disparate observation qualities of different viewpoints.

Parallel to this underutilization of geometry is the failure to allocate computation by information density.
In occupancy prediction, information density drops sharply as resolution increases: the coarse grid (${\sim}10$K voxels) contains semantically meaningful content at nearly every location, while the fine grid (${\sim}640$K voxels) is predominantly empty.
Dense methods apply uniform computation across all scales, with U-Net skip connections dominating the decoder cost; under matched dimensions a SurroundOcc-style decoder reaches $\sim$73.4G FLOPs for a single scale transition, most spent on 3D convolutions over empty voxels.
Sparse methods like GaussianFormer-2~\cite{gaussianformer2} reduce computation by using fewer primitives, but Gaussian-to-voxel splatting loses geometric detail.
Both paradigms pay a price: dense methods waste compute on empty space, while sparse methods sacrifice completeness.

\begin{figure}[t]
  \centering
  \includegraphics[width=\linewidth]{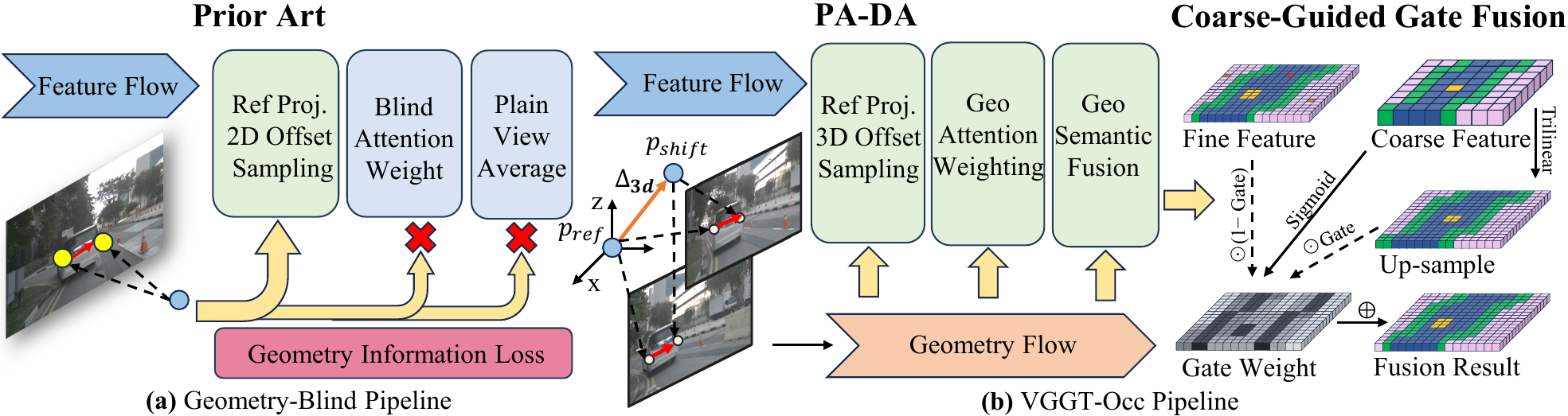}
  \caption{\textbf{VGGT-Occ overview.}
(a)~Prior methods restrict camera geometry to initial projection, leaving subsequent attention stages geometry-blind.
(b)~VGGT-Occ injects projection geometry into all attention stages via PA-DA, and allocates computation by voxel density via coarse-guided gated fusion.}
  \label{fig:overview}
\end{figure}

In this paper, we propose \textbf{VGGT-Occ}, built on two design principles that address the aforementioned limitations (Figure~\ref{fig:overview}) and incorporate a 3D foundation model into 3D occupancy map reconstruction.
\emph{Principle~(i) Geometric Ubiquity}: 3D geometric information should be embedded throughout the entire pipeline, from the multi-view encoder to the attention-based occupancy head.
\emph{Principle~(ii) Density-Awareness}: Computational budget should be allocated by information density, concentrating expensive operations where they yield the maximum semantic return.

To instantiate principle~(i), we unify the encoding process using a geometry-grounded Transformer (VGGT~\cite{vggt}), enabling cross-view geometry reasoning and feature alignment inherently during the encoding phase, rather than via post-hoc fusion.
Within the occupancy head, we introduce \textbf{PA-DA (Projection-Aware Deformable Attention)}, a mechanism that explicitly injects camera geometry into all three stages of cross-attention:
\textbf{Stage~1} learns true 3D offsets and projects them to each camera's 2D plane, ensuring cross-view consistency and distance-adaptive sampling.
\textbf{Stage~2} uses the projection Jacobian to encode per-point observation quality as an additive attention bias, automatically suppressing severely distorted sampling directions via a logarithmic penalty.
\textbf{Stage~3} fuses cross-camera features through a view-quality semantic gate, where observation quality and feature semantics jointly determine the fusion weights.
Together, VGGT and PA-DA ensure that geometric constraints persist from the initial feature extraction to the final 3D lifting.

In pursuit of principle~(ii), we introduce a \textbf{Density-Aware Computation Allocation} strategy.
The occupancy head is sequentially structured: a single embedding at the coarsest scale is refined through successive Patch Splitting layers, with each scale's output feeding into the next as query initialization.
This ensures that coarse-scale features carry structurally meaningful semantics, enabling the gating mechanism to make informed fusion decisions at subsequent scales.
The expensive 2D-to-3D lifting (cross-attention) is restricted to information-dense coarse scales ($10$K and $80$K voxels),  whereas the fine-grained scale ($640$K voxels) is processed by lightweight local depthwise convolutions (dim$=$64).
To bridge these scales, we propose \textbf{Coarse-Guided Gated Fusion}: channel-wise gates are computed at low resolution using coarse semantic features to determine feature criticality, upsampled for high-resolution execution, and smoothed by depthwise 3D convolutions.
This strategy achieves an ${\bf 8.9\times}$ reduction in fusion cost (~73.4G to ~8.2G FLOPs for the S1$\to$S2 transition) while preserving full geometric coverage, embodying a sparse computation philosophy within a dense framework.

Our main contributions are:
\begin{enumerate}
  \item We propose \textbf{PA-DA}, a Projection-Aware Deformable Attention that injects camera geometry into all three stages of cross-attention (3D offset projection, Jacobian-guided attention weighting, and view-quality semantic gated fusion). Ablation studies confirm the independent contribution of each stage.
  \item We design a \textbf{sequential coarse-to-fine decoding} paradigm with \textbf{density-aware gated fusion}: cross-attention is allocated to information-dense coarse scales while fine scales use only lightweight convolutions; channel-wise gates bridge scales with substantially lower fusion cost while maintaining strong performance.
  \item VGGT-Occ achieves state-of-the-art results on SurroundOcc-nuScenes: 33.00\% IoU and 21.08\% mIoU, outperforming GaussianFormer-2 and QuadricFormer, with only ${\sim}41$M trainable parameters in the occupancy head.
\end{enumerate}

\section{Related Work}

\paragraph{Geometry in Multi-View 3D Perception.}
Camera geometry is fundamental to 2D-to-3D correspondence, yet the depth of its exploitation varies significantly. The 2D-to-BEV transformation originated with OFT~\cite{oft} and LSS~\cite{lss} and was later adopted by BEVDet~\cite{bevdet}. Most subsequent methods~\cite{bevformer,surroundocc,occformer,li2023fbocc,wang2024panoocc,petr} use geometry solely to determine reference points for 2D feature indexing; offset learning, attention weighting, and cross-camera aggregation remain purely data-driven. Sparse methods go further: GaussianFormer-2~\cite{gaussianformer2} and QuadricFormer~\cite{quadricformer} use ray-based depth for primitive initialization, while Sparse4D~\cite{sparse4d} and Far3D~\cite{jiang2024far3d} learn 3D offsets, but none modulate attention weights or aggregation via projection geometry. Recent advances in monocular depth~\cite{depthanything} and 3D foundation models~\cite{vggt,unipr-3d,relocvggt} have enabled stronger geometric priors for occupancy~\cite{zhu2026drocc}. VGGT-Occ leverages such a foundation model as its encoder and further injects projection geometry into every attention stage, pushing geometric reasoning beyond the initial projection.

\paragraph{Deformable Attention for 3D Perception.}
Deformable DETR~\cite{deformabledetr} introduced sparse sampling with learned 2D offsets; DETR3D~\cite{detr3d} pioneered the 3D-to-2D query paradigm for multi-view detection, later adapted by BEVFormer~\cite{bevformer} for 3D perception. Sparse4D~\cite{sparse4d} and Far3D~\cite{jiang2024far3d} learn 3D offsets for cross-view consistency, yet without leveraging projection geometry for attention weighting or aggregation. GaussianFormer3D~\cite{zhao2026gaussianformer3d} uses 3D deformable attention in LiDAR-camera fused space, orthogonal to our camera-only setting.
PA-DA extends this line by injecting geometry into all three attention stages: 3D offset projection, Jacobian-modulated bias, and view-quality gated fusion.

\paragraph{3D Occupancy Prediction.}
Early dense methods~\cite{monoscene,surroundocc,occformer} pioneered multi-scale voxel architectures with heavy 3D convolution decoders; subsequent works~\cite{li2023fbocc,wang2024panoocc,ctfocc,oh2025protoocc,shi2026bepo} extended this with improved view transformation and dual-branch designs. FlashOcc~\cite{flashocc} accelerates prediction via a channel-to-height plugin, while RenderOcc~\cite{pan2024renderocc} and GaussRender~\cite{gaussrender} add rendering-based supervision. These methods achieve complete coverage yet apply uniform computation across all scales, including empty fine-scale voxels.
Sparse methods~\cite{gaussianformer,gaussianformer2,liu2024sparseocc,zhang2025sqs,quadricformer,zuo2025gaussianworld} represent scenes with a small set of 3D Gaussians~\cite{3dgs} or superquadrics for efficient computation, embodying an ``information-density-driven'' philosophy, but sacrifice geometric detail through splatting. SelfOcc~\cite{huang2024selfocc} explores self-supervision via neural rendering on dense voxel grids. Cam4DOcc~\cite{ma2024cam4docc} extends occupancy to 4D forecasting.
VGGT-Occ reconciles both paradigms: dense voxel grid for complete coverage, with sparse computation philosophy applied within it (cross-attention only at coarse scales, lightweight convolutions at the fine scale).

\section{Method}

\subsection{Overview}

\begin{figure}[t]
  \centering
  \includegraphics[width=\linewidth]{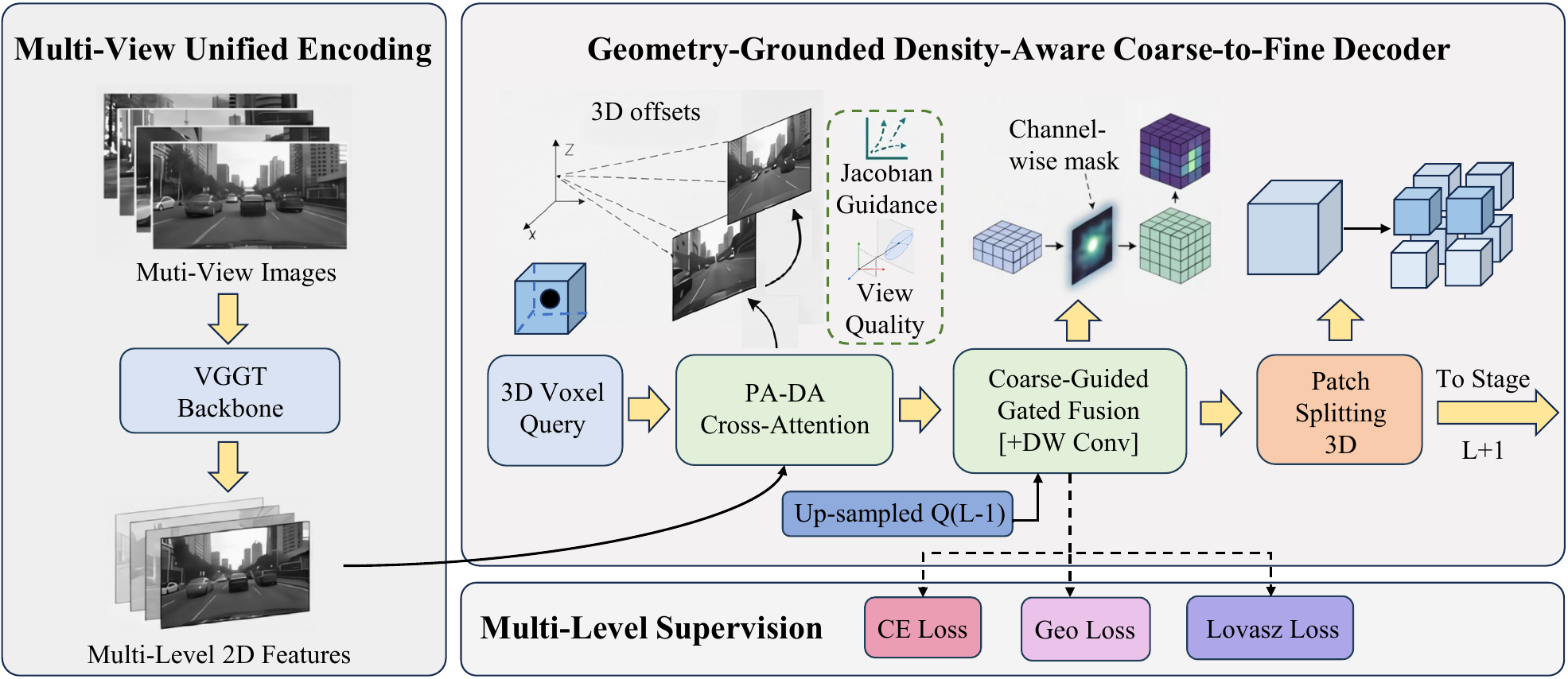}
  \caption{\textbf{VGGT-Occ architecture.} VGGT unified encoding produces multi-scale 2D features. PA-DA injects projection geometry into three stages of cross-attention at coarse scales. Density-aware decoder uses convolutions only at fine scale, with coarse-guided gated fusion bridging scales.}
  \label{fig:architecture}
\end{figure}

VGGT-Occ instantiates two design principles: \emph{(i)} geometric reasoning should permeate the pipeline from the multi-view encoder to the attention-based head, and \emph{(ii)} computation should be allocated by information density.
Figure~\ref{fig:architecture} illustrates the architecture.
All six camera views are encoded jointly by VGGT~\cite{vggt}, a geometry-grounded Transformer that performs cross-view reasoning during encoding.
The occupancy head operates across three spatial scales ($S_0$: $50{\times}50{\times}4$, $S_1$: $100{\times}100{\times}8$, $S_2$: $200{\times}200{\times}16$) in a \emph{sequential} manner.
A learnable embedding grid at $S_0$ is refined through successive Patch Splitting layers to form a \emph{coarse-to-fine} structure, with each scale's output initializing the queries of the next rather than predicting independently.
Coarse scales ($S_0$, $S_1$) apply PA-DA cross-attention (Section~\ref{sec:pada}) to lift 2D features into 3D, while the fine scale ($S_2$) uses only lightweight convolutions and receives coarse features via gated fusion (Section~\ref{sec:fusion}).

\FloatBarrier

\subsection{PA-DA: Projection-Aware Deformable Attention}
\label{sec:pada}

\begin{figure}[t]
  \centering
  \includegraphics[width=\linewidth]{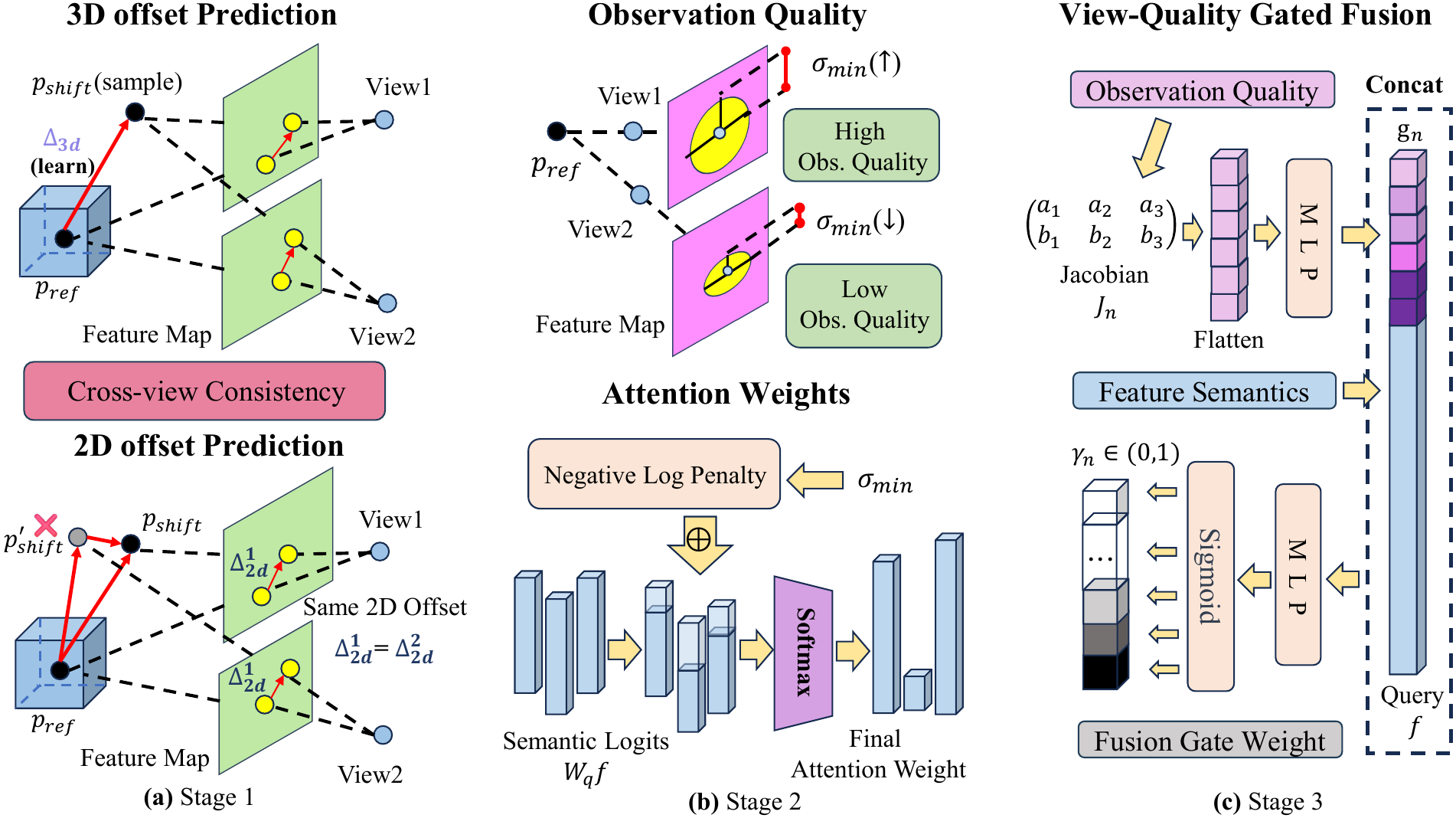}
  \caption{\textbf{PA-DA: three-stage projection-aware deformable attention.} Stage~1 learns 3D offsets and projects them to each camera's image plane for cross-view consistency. Stage~2 decomposes the projection Jacobian to extract $\sigma_{\min}$, encoding per-point observation quality as an additive log-bias. Stage~3 embeds the full $2{\times}3$ Jacobian for per-camera, per-channel gated fusion.}
  \label{fig:pada}
\end{figure}

We first review the standard deformable cross-attention~\cite{deformabledetr} used in multi-view 3D perception.
Given a 3D voxel query $\mathbf{q}$, its 3D reference point $\mathbf{p}_{\mathrm{ref}}$ is projected onto the 2D feature map of each camera $n$ via the pinhole model to obtain a reference location $(u_n^0, v_n^0)$.
Multiple sampling points are then generated around each reference location by learning 2D offsets, and the sampled features are aggregated with attention weights computed purely from the query.
In this pipeline, camera intrinsics $\mathbf{K}_n$ and extrinsics $\mathbf{E}_n$ are used \emph{only} at the initial projection. The subsequent offset learning, attention weighting, and cross-camera aggregation are entirely data-driven.
PA-DA closes this gap by injecting projection geometry into all three stages (Figure~\ref{fig:pada}).

\subsubsection{3D Offset Learning with Projection}
\label{sec:stage1}

To resolve the cross-view inconsistency of 2D offset learning, we predict independent 3D offsets $\boldsymbol{\Delta}^{(h,l,k)}_{3d} \in \mathbb{R}^3$ (one per attention head $h$, feature level $l$, and sampling point $k$) from the query feature via a small MLP and project each shifted point to every camera:
\begin{equation}
  \mathbf{p}^{(h,l,k)}_{\mathrm{shift}} = \mathbf{p}_{\mathrm{ref}} + \boldsymbol{\Delta}^{(h,l,k)}_{3d}, \qquad
  (u^{(h,l,k)}_n, v^{(h,l,k)}_n) = \pi(\mathbf{p}^{(h,l,k)}_{\mathrm{shift}};\, \mathbf{K}_n, \mathbf{E}_n),
  \label{eq:project}
\end{equation}
where $\pi$ is the standard pinhole projection. All sampling points across heads, levels, and cameras are geometrically consistent with the same 3D reference frame.
Because the offset lives in 3D, the same semantic direction produces geometrically correct 2D shifts in every camera, ensuring cross-view consistency.
Moreover, the same 3D offset yields a large 2D displacement for nearby cameras and a small one for distant cameras, providing natural distance-adaptive sampling.

\subsubsection{Jacobian-Guided Attention Weighting}
\label{sec:stage2}

The projection Jacobian $\mathbf{J} = \frac{\partial(u,v)}{\partial(X,Y,Z)} \in \mathbb{R}^{2\times3}$ is rank-2, with its null space along the viewing ray. Its smaller non-zero singular value $\sigma_{\min}$ quantifies the weakest observable spatial direction, a conservative measure of observation quality. We compute $\sigma_{\min}$ at each shifted 3D point from Eq.~\eqref{eq:project} by reusing the projection's intermediate Jacobian with negligible overhead.
For a pinhole camera $\sigma_{\min}$ simplifies to $f / (Z \cdot W)$ in normalized image coordinates, acting as a depth penalty; in a multi-camera setup this already differentiates viewing quality since oblique cameras exhibit larger $Z$ to the same 3D point.
Importantly, the SVD formulation is \emph{projection-agnostic}: it applies without modification to non-pinhole cameras (e.g., fisheye, panoramic) where a closed-form observation quality measure is unavailable, making the approach portable across camera models.

This observability is injected as an additive bias into the attention logits, resolving the blindness to per-point observation quality:
\begin{equation}
  \mathrm{obs} = \sigma_{\min}(\mathbf{J}_{\mathrm{shift}}).\mathrm{detach}(), \qquad
  \mathbf{a} = \mathrm{softmax}\!\big(\mathbf{W}_q \mathbf{f} + s_{h,l} \cdot \log(\mathrm{obs} + \epsilon)\big),
  \label{eq:jacobian_attn}
\end{equation}
where $\mathbf{f}$ is the query feature, $\mathbf{W}_q$ the learnable projection, and $s_{h,l}$ a per-head, per-level learnable scalar initialized to $0.1$.
The Jacobian is computed in normalized image coordinates (divided by image width and height), so $\sigma_{\min}$ is dimensionless.
The small initialization lets the network gradually discover the geometric prior.
We detach the gradient of the observability term, so the 3D offset parameters learn only from the task loss and are not biased toward ``seeking high observability.''

\paragraph{Log Penalty.}
Since $\sigma_{\min} \in (0, {\sim}0.2)$ in normalized coordinates, $\log(\mathrm{obs} + \epsilon)$ produces a negative penalty that suppresses low-quality sampling points via softmax.
As a result, frontal/nearby points receive small penalties and high weights, while oblique/distant points are automatically down-weighted.

\subsubsection{View-Quality Semantic Gated Fusion}
\label{sec:stage3}

PA-DA replaces uniform cross-camera averaging with per-camera, per-channel gated fusion. It leverages the full $2{\times}3$ Jacobian matrix (6 dimensions) to predict fusion weights that reflect each view's observation quality.
Unlike Stage~2, which uses only $\sigma_{\min}$ (a scalar), the full Jacobian preserves directional information about which spatial directions are well or poorly observed.

For each camera $n$, we embed its Jacobian via a small 2-layer MLP and combine with the query to produce per-channel fusion weights:
\begin{align}
  \mathbf{g}_n &= \mathrm{MLP}_{\mathrm{geo}}\!\big(\mathrm{flatten}(\mathbf{J}_n) \times 1000\big) \in \mathbb{R}^{C}, \label{eq:geo_embed} \\
  \boldsymbol{\gamma}_n &= \mathrm{MLP}_{\mathrm{gate}}\!([\,\mathbf{f} \;\|\; \mathbf{g}_n\,]) \in \mathbb{R}^{C}, \label{eq:gate}
\end{align}
where $[\cdot\|\cdot]$ denotes channel-wise concatenation. Full MLP specifications are given in Appendix~\ref{sec:pada_details}.
The fused output aggregates across cameras:
\begin{equation}
  \mathbf{o} = \frac{\sum_n \boldsymbol{\gamma}_n \odot \mathbf{v}_n}{\sum_n \boldsymbol{\gamma}_n},
  \label{eq:fusion}
\end{equation}
where $\mathbf{v}_n$ is the sampled feature from camera $n$ and $\odot$ denotes element-wise multiplication. The Jacobian is gradient-detached so it does not bias offset learning.

\subsection{Density-aware coarse-guided gated fusion}
\label{sec:fusion}

Information density in occupancy prediction drops sharply with resolution: the coarse grid (${\sim}10$K voxels) is semantically rich, while the fine grid (${\sim}640$K voxels) is predominantly empty.
We therefore restrict expensive PA-DA cross-attention to the coarse scales and use only lightweight 3D convolutions at the fine scale (Figure~\ref{fig:architecture}).

To propagate coarse-scale semantics to the fine scale, we employ a coarse-guided gated fusion.
The key idea is that high-resolution features do not require high-resolution fusion decisions: thanks to the sequential architecture, coarse-scale features already contain rich semantic context to inform fusion decisions at the fine scale.
We compute channel-wise gates at low resolution and upsample them for high-resolution execution:
\begin{align}
  \mathbf{g} &= \sigma\!\big(\mathrm{MLP}(\mathbf{h}_{\mathrm{prev}})\big), \label{eq:cf_gate} \\
  \hat{\mathbf{g}} &= \mathrm{Up}(\mathbf{g}), \qquad
  \hat{\mathbf{h}}_{\mathrm{prev}} = \mathrm{Up}\!\big(\mathrm{Proj}(\mathbf{h}_{\mathrm{prev}})\big), \label{eq:cf_up} \\
  \mathbf{h}_{\mathrm{fused}} &= \hat{\mathbf{g}} \odot \hat{\mathbf{h}}_{\mathrm{prev}} + (1 - \hat{\mathbf{g}}) \odot \mathbf{h}_{\mathrm{curr}}, \label{eq:cf_fuse}
\end{align}
where $\mathbf{h}_{\mathrm{prev}}$ is the previous scale's output, $\mathrm{Proj}$ linearly projects to the target dimension when needed, and $\mathrm{Up}$ denotes trilinear upsampling.
The projection is applied before upsampling to reduce the cost of the interpolation.

Upsampling the gate from low to high resolution can introduce boundary artifacts.
A depthwise 3D convolution~\cite{convnext} smooths the fused features with minimal overhead:
\begin{equation}
  \mathbf{h}_{\mathrm{out}} = \mathrm{LayerNorm}\!\big(\mathrm{DW\text{-}Conv3d}(\mathbf{h}_{\mathrm{fused}})\big),
  \label{eq:dwconv}
\end{equation}
where the convolution uses kernel size $3$ with groups equal to the channel dimension.
The entire fusion pipeline reduces the cost from $\sim$73.4G FLOPs (SurroundOcc-style decoder) to $\sim$8.2G FLOPs, an $8.9\times$ reduction.





\section{Experiments}
\label{sec:experiments}

\subsection{Setup}

\textbf{Datasets.}
We evaluate on SurroundOcc-nuScenes~\cite{surroundocc} (700 training / 150 validation scenes), built upon the nuScenes dataset~\cite{nuscenes}. Other 3D occupancy benchmarks include SSCBench~\cite{sscbench}, which unifies multiple datasets under a common semantic scene completion protocol.
Both define a voxel grid of $200 {\times} 200 {\times} 16$ covering $100\text{m} \times 100\text{m} \times 8\text{m}$ at 0.5m resolution, with 17 semantic classes (1 free + 16 categories).
Following standard protocol~\cite{surroundocc}, we report two metrics:
\begin{equation}
\mathrm{IoU} = \frac{\mathrm{TP}}{\mathrm{TP} + \mathrm{FP} + \mathrm{FN}}, \qquad
\mathrm{mIoU} = \frac{1}{C} \sum_{c=1}^{C} \frac{\mathrm{TP}_c}{\mathrm{TP}_c + \mathrm{FP}_c + \mathrm{FN}_c},
\end{equation}
where TP, FP, FN denote true positive, false positive, and false negative voxel predictions; $C{=}16$ is the number of semantic classes.
IoU measures geometry completion (occupied vs.\ free), and mIoU averages per-class semantic IoU.

\textbf{Architecture.}
We use VGGT~\cite{vggt} (1.2B params, 24 alternating attention blocks, DINOv2 ViT-L~\cite{dinov2,vit} as patch tokenizer) as the frozen multi-view encoder.
Input images are $378 {\times} 672$ across all 6 surround cameras.
The occupancy head has 3 scales: $50 {\times} 50 {\times} 4$, $100 {\times} 100 {\times} 8$, $200 {\times} 200 {\times} 16$.
Coarse scales use PA-DA cross-attention + depthwise separable 3D Conv (ConvNeXt-style~\cite{convnext}); the fine scale uses convolutions only.
PA-DA uses $n_\text{heads}{=}4$, $n_\text{points}{=}4$.
The head has $\sim$41M trainable parameters.
Full hyperparameters are provided in Appendix~\ref{sec:impl_details}.

\textbf{Training.}
We train 20 epochs with AdamW~\cite{adamw}, learning rate $10^{-4}$, weight decay 0.01, cosine decay to $10^{-6}$, effective batch size 36 (3 GPU $\times$ BS3 $\times$ 4 accum), on 3$\times$ RTX 4090 (48G)\footnote{The RTX 4090 is a modified version with 48\,GB VRAM; the standard consumer variant has 24\,GB.}.
Backbone is frozen.
Loss: CE + semantic scene completion~\cite{surroundocc} + geometry scene completion~\cite{surroundocc} + Lov\'{a}sz-Softmax~\cite{lovasz}.
Further details (warmup, gradient clipping, label smoothing, EMA, augmentations, loss equations) are in Appendix~\ref{sec:impl_details}.

\subsection{Main Results}

\begin{table}[t]
\centering
\scriptsize
\caption{\textbf{3D semantic occupancy on SurroundOcc-nuScenes.}}
\label{tab:main}
\setlength{\tabcolsep}{1.2pt}
\begin{tabular}{@{}l|cc|*{16}{c}@{}}
\toprule
Method & IoU & mIoU & \rotatebox{90}{barrier} & \rotatebox{90}{bicycle} & \rotatebox{90}{bus} & \rotatebox{90}{car} & \rotatebox{90}{const.\ veh.} & \rotatebox{90}{motorcycle} & \rotatebox{90}{pedestrian} & \rotatebox{90}{traffic\ cone} & \rotatebox{90}{trailer} & \rotatebox{90}{truck} & \rotatebox{90}{drive.\ surf.} & \rotatebox{90}{other\ flat} & \rotatebox{90}{sidewalk} & \rotatebox{90}{terrain} & \rotatebox{90}{manmade} & \rotatebox{90}{vegetation} \\
& & & \textcolor{clr_barrier}{\rule{5pt}{5pt}} & \textcolor{clr_bicycle}{\rule{5pt}{5pt}} & \textcolor{clr_bus}{\rule{5pt}{5pt}} & \textcolor{clr_car}{\rule{5pt}{5pt}} & \textcolor{clr_cveh}{\rule{5pt}{5pt}} & \textcolor{clr_motor}{\rule{5pt}{5pt}} & \textcolor{clr_ped}{\rule{5pt}{5pt}} & \textcolor{clr_tcone}{\rule{5pt}{5pt}} & \textcolor{clr_trailer}{\rule{5pt}{5pt}} & \textcolor{clr_truck}{\rule{5pt}{5pt}} & \textcolor{clr_drvsurf}{\rule{5pt}{5pt}} & \textcolor{clr_othflat}{\rule{5pt}{5pt}} & \textcolor{clr_sidewlk}{\rule{5pt}{5pt}} & \textcolor{clr_terrain}{\rule{5pt}{5pt}} & \textcolor{clr_manmade}{\rule{5pt}{5pt}} & \textcolor{clr_veg}{\rule{5pt}{5pt}} \\
\midrule
MonoScene~\cite{monoscene} & 23.96 & 7.31 & 4.03 & 0.35 & 8.00 & 8.04 & 2.90 & 0.28 & 1.16 & 0.67 & 4.01 & 4.35 & 27.72 & 5.20 & 15.13 & 11.29 & 9.03 & 14.86 \\
BEVFormer~\cite{bevformer} & 30.50 & 16.75 & 14.22 & 6.58 & 23.46 & 28.28 & 8.66 & 10.77 & 6.64 & 4.05 & 11.20 & 17.78 & 37.28 & 18.00 & 22.88 & 22.17 & 13.80 & 22.21 \\
TPVFormer~\cite{tpvformer} & 30.86 & 17.10 & 15.96 & 5.31 & 23.86 & 27.32 & 9.79 & 8.74 & 7.09 & 5.20 & 10.97 & 19.22 & 38.87 & 21.25 & 24.26 & 23.15 & 11.73 & 20.81 \\
OccFormer~\cite{occformer} & 31.39 & 19.03 & 18.65 & 10.41 & 23.92 & 30.29 & 10.31 & 14.19 & 13.59 & 10.13 & 12.49 & 20.77 & 38.78 & 19.79 & 24.19 & 22.21 & 13.48 & 21.35 \\
SurroundOcc~\cite{surroundocc} & 31.49 & 20.30 & 20.59 & 11.68 & 28.06 & 30.86 & 10.70 & 15.14 & 14.09 & 12.06 & \textbf{14.38} & 22.26 & 37.29 & 23.70 & 24.49 & 22.77 & 14.89 & 21.86 \\
GaussianFormer-2~\cite{gaussianformer2} & 31.74 & 20.82 & \textbf{21.39} & 13.44 & \textbf{28.49} & 30.82 & 10.92 & 15.84 & 13.55 & 10.53 & 14.04 & 22.92 & 40.61 & 24.36 & 26.08 & 24.27 & 13.83 & 21.98 \\
GaussRender~\cite{gaussrender} & 32.61 & 20.82 & 20.32 & 13.22 & 28.32 & \textbf{31.05} & 10.92 & 15.65 & 12.84 & 8.91 & 13.29 & 22.76 & \textbf{41.22} & \textbf{24.48} & \textbf{26.38} & \textbf{25.20} & 15.31 & 23.25 \\
QuadricFormer~\cite{quadricformer} & 31.22 & 20.12 & 19.58 & 13.11 & 27.27 & 29.64 & \textbf{11.25} & 16.26 & 12.65 & 9.15 & 12.51 & 21.24 & 40.20 & 24.34 & 25.69 & 24.24 & 12.95 & 21.86 \\
\midrule
\textbf{VGGT-Occ} ($T{=}1$) & \textbf{33.00} & \textbf{21.08} & 19.85 & \textbf{16.49} & 27.14 & 28.95 & 9.37 & \textbf{18.19} & \textbf{15.28} & \textbf{14.31} & 12.55 & \textbf{23.03} & 38.77 & 24.37 & 25.64 & 24.51 & \textbf{15.41} & \textbf{23.39} \\
\textbf{VGGT-Occ} ($T{=}2$) & \textbf{33.64} & \textbf{21.43} & 20.49 & 16.51 & 26.88 & 27.87 & 9.81 & 17.97 & 15.72 & 15.07 & 13.08 & 22.94 & 39.50 & 25.53 & 26.49 & 25.09 & 15.84 & 24.11 \\
\bottomrule
\end{tabular}
\end{table}

Table~\ref{tab:main} compares VGGT-Occ with camera-only methods on SurroundOcc-nuScenes.
VGGT-Occ with $T{=}1$ temporal frame (each comprising 6 surround images) achieves 33.00\% IoU and 21.08\% mIoU, surpassing GaussianFormer-2 (20.82\% mIoU) and QuadricFormer (20.12\% mIoU), despite using a $378 {\times} 672$ input resolution, substantially lower than the $900 {\times} 1600$ resolution employed by SurroundOcc, GaussianFormer-2, and QuadricFormer.
VGGT-Occ uses a 1.2B frozen backbone while the compared baselines employ substantially smaller encoders (e.g., ResNet-101, $\sim$45M). Despite this disparity, Table~\ref{tab:efficiency} shows that VGGT-Occ is $1.6\times$ faster at inference than GaussianFormer-2 with 42\% less GPU memory, confirming that the large backbone does not preclude efficient deployment.
The model is trained exclusively on single temporal frames, yet VGGT's unified encoding naturally accepts multi-frame input at inference: simply feeding $T{=}2$ temporal steps through the frozen encoder boosts performance to 33.64\% IoU and 21.43\% mIoU, with no additional training or architectural changes.
Beyond aggregate metrics, VGGT-Occ shows particular strength on categories where geometric reasoning matters most: small objects: pedestrian (15.28\%), bicycle (16.49\%), traffic cone (14.31\%), motorcycle (18.19\%) benefit from Jacobian-guided attention that preserves fine-grained sampling in frontal views while suppressing noisy observations from oblique cameras.
Compared to QuadricFormer, which achieves efficiency through a sparse superquadric representation, VGGT-Occ reaches higher accuracy through a fundamentally different route, deeper geometric reasoning within a dense voxel framework, with only $\sim$41M trainable head parameters.
The progression from BEVFormer (16.75) through SurroundOcc (20.30) to VGGT-Occ (21.08, $T{=}1$) tracks with the depth of geometric information utilization, consistent with our core thesis.

\begin{wraptable}{r}{0.48\textwidth}
\centering
\footnotesize
\caption{\textbf{Inference efficiency.} Measured on a single RTX 4090 (48G), batch size~$=$~1.}
\label{tab:efficiency}
\begin{tabular}{@{}lcc@{}}
\toprule
Method & Latency (ms) & Memory (MB) \\
\midrule
SurroundOcc & 343.7 & 5,733 \\
GaussianFormer-2 & 747.7 & 23,847 \\
VGGT-Occ & 483.2 & 13,887 \\
\bottomrule
\end{tabular}
\end{wraptable}
\textbf{Efficiency.} The frozen VGGT backbone (1.2B) dominates total parameters, but training requires gradients only for the $\sim$41M head parameters ($<$3\% of total).
Our S1$\to$S2 fusion costs $\sim$8.2G FLOPs vs.\ $\sim$73.4G for the SurroundOcc-style decoder, an $8.9\times$ reduction at matching accuracy (Table~\ref{tab:abl_fusion}).
Table~\ref{tab:efficiency} reports inference latency and GPU memory. VGGT-Occ's 483.2ms breaks down into 241.2ms (VGGT backbone) + 242.0ms (OccHead).
VGGT-Occ's speed advantage partly stems from its lower input resolution ($378{\times}672$ vs.\ $900{\times}1600$ for all baselines); despite the 1.2B frozen encoder, it is $1.6\times$ faster than GaussianFormer-2 (747.7ms) with 42\% less memory.
For GaussianFormer-2, the network inference uses only $\sim$4GB, but Gaussian-to-voxel splatting dominates total memory.
All memory figures are measured via \texttt{nvidia-smi}, capturing CUDA context and cuDNN workspace that PyTorch's \texttt{max\_memory\_allocated} does not track.
The large frozen backbone lets the lightweight head focus on geometric 2D-to-3D lifting, while backbone features can be shared across downstream tasks. Table~\ref{tab:abl_backbone_head} confirms the head is backbone-agnostic: PA-DA improves accuracy even with DINOv2-Base.

\begin{figure}[t]
  \centering
  \includegraphics[width=\linewidth]{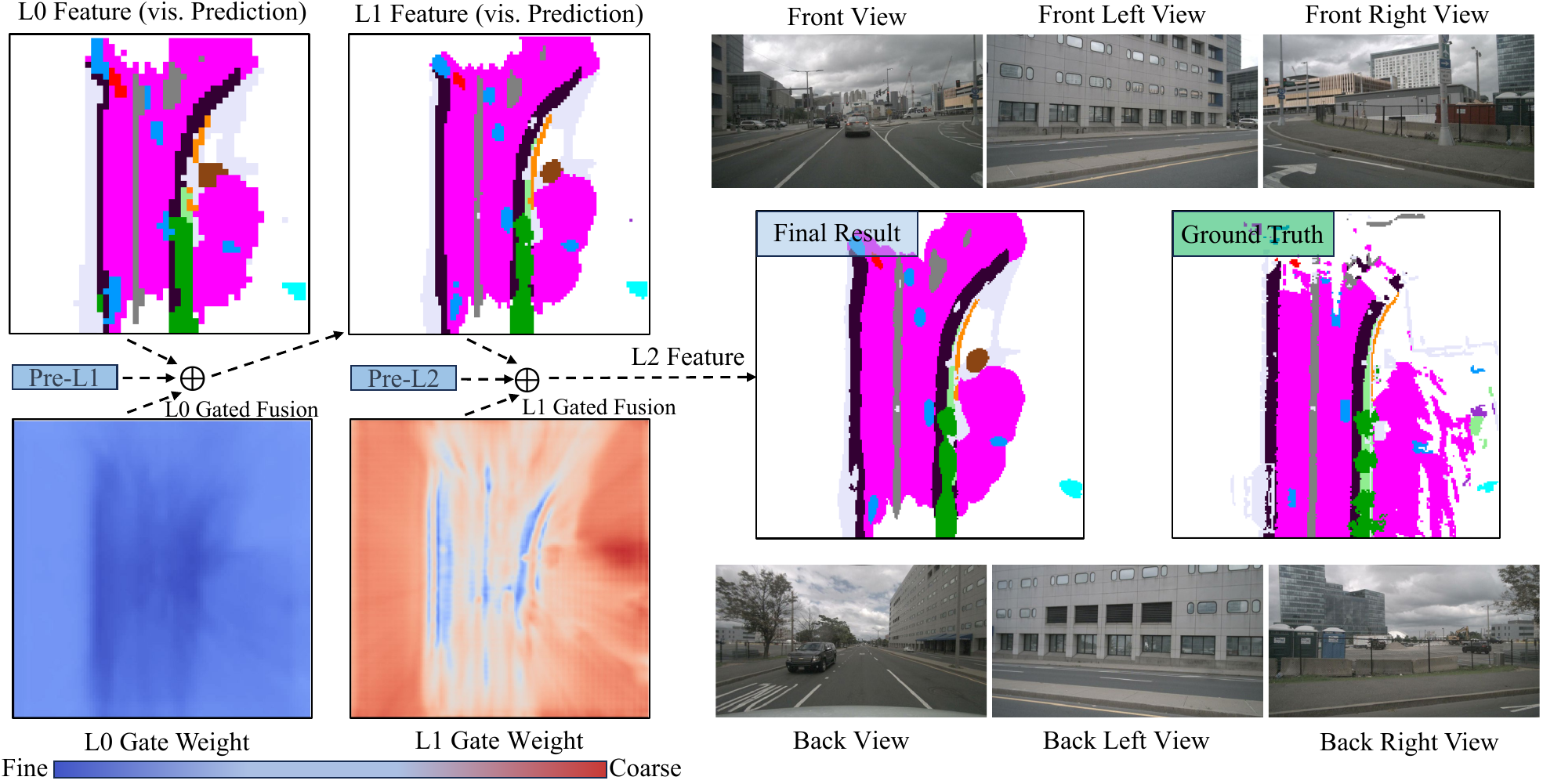}
  \caption{\textbf{Visualization of the coarse-to-fine gated fusion.}
  (Left) Cascaded fusion pipeline: base features ($L_0$, $L_1$) fused with intermediate predictions (Pre-$L_1$, Pre-$L_2$) via learned gates.
  (Right) Multi-view RGB inputs, final prediction ($L_2$), and ground truth. Warmer gate colors indicate stronger coarse-level reliance.}
  \label{fig:gate}
\end{figure}

\paragraph{Gating Behavior.}
Figure~\ref{fig:gate} visualizes the learned coarse-to-fine gating at both fusion levels (warmer colors indicate stronger reliance on coarse features).
The $L_0$ gate predominantly trusts fine-scale features, since $L_1$ itself benefits from PA-DA cross-view attention.
The $L_1$ gate relies more on coarse semantics (as $L_2$ has no cross-view attention), but defers to fine-scale features at object boundaries where spatial precision is required.
This level-adaptive behavior emerges without explicit supervision, confirming that coarse-guided gating learns physically meaningful and architecturally appropriate modulation.

\FloatBarrier
\subsection{Ablation studies}

All ablations use the same VGGT backbone and training recipe (Appendix~\ref{sec:impl_details}), varying only the component under study.

\subsubsection{Progressive Geometry Injection in PA-DA}

Each stage targets a distinct source of geometric blindness in standard cross-attention.

\begin{wraptable}{r}{0.50\textwidth}
\centering
\caption{\textbf{PA-DA stage-wise ablation.}}
\label{tab:abl_pada}
\begin{tabular}{@{}lccccl@{}}
\toprule
Configuration & S1 & S2 & S3 & IoU & mIoU \\
\midrule
Baseline & & & & 31.84 & 20.21 \\
+ Stage~1 & \checkmark & & & 32.45 & 20.69 \\
+ S1 + 2 & \checkmark & \checkmark & & 32.84 & 20.87 \\
\textbf{+ S1 + 2 + 3} & \checkmark & \checkmark & \checkmark & \textbf{33.00} & \textbf{21.08} \\
\bottomrule
\end{tabular}
\end{wraptable}

Table~\ref{tab:abl_pada} reports the core ablation: starting from a baseline without geometric injection (2D offsets, simple cross-camera averaging) and progressively activating PA-DA stages.
Stage~1 (3D offset learning, projected to each camera) provides cross-view consistency and distance-adaptive sampling: a ``look above'' offset maps to correct 2D directions in all six views, and near-camera offsets produce larger 2D displacements than far-camera ones.
Stage~2 (Jacobian-guided attention) penalizes sampling points with low $\sigma_{\min}$, which for pinhole cameras reflects large depth; in a multi-camera setup this selectively suppresses features from oblique and distant cameras while retaining high-quality frontal observations.
Stage~3 (semantic gated fusion) uses the full $2{\times}3$ projection Jacobian to predict per-camera per-channel fusion weights, letting different feature channels trust different cameras based on their geometric observation quality.
Each stage independently improves both IoU and mIoU, and the three stages compound, establishing a causal chain between geometric injection and occupancy prediction accuracy.

\subsubsection{Gated Fusion, Backbone, and Head Contributions}

\begin{wraptable}{r}{0.45\textwidth}
\centering
\footnotesize
\caption{\textbf{Gated fusion design.}}
\label{tab:abl_fusion}
\begin{tabular}{@{}lccc@{}}
\toprule
Fusion & FLOPs (G) & IoU & mIoU \\
\midrule
U-Net (SurroundOcc) & $\sim$73.4 & 32.95 & 21.10 \\
\midrule
No fusion & 0 & 30.79 & 19.87 \\
Direct add & negl. & 31.68 & 20.48 \\
Scalar gate & 2.68 & 32.86 & 20.83 \\
Channel gate & 6.02 & 32.97 & 21.02 \\
\textbf{Channel gate + DW} & \textbf{$\sim$8.2} & \textbf{33.00} & \textbf{21.08} \\
\bottomrule
\end{tabular}
\end{wraptable}

Table~\ref{tab:abl_fusion} ablates the S1$\to$S2 fusion design (100$\times$100$\times$8 to 200$\times$200$\times$16, 256 to 64 channels).
The U-Net baseline represents the SurroundOcc-style decoder: deconvolution (kernel 2$^3$) upsamples the coarse feature, followed by addition and a 3D convolution (kernel 3$^3$).
Our variants all use trilinear upsampling ($\sim$0 FLOPs) and share a 256$\to$64 channel MLP.
\textbf{No fusion} omits cross-scale connection entirely (lower bound).
\textbf{Direct add} upsamples and adds coarse features without learned modulation.
\textbf{Scalar gate} learns a per-voxel scalar weight (256$\to$1 MLP + sigmoid).
\textbf{Channel gate} learns per-channel weights (256$\to$64 MLP + sigmoid), enabling independent modulation of each feature dimension.
\textbf{Channel gate + DW}, our final design (Eq.~\ref{eq:cf_fuse}--\ref{eq:dwconv}), appends a depthwise separable 3D convolution (kernel 3$^3$, groups$=$channels) to smooth upsampling artifacts.
The progression shows that learned gating is essential and per-channel modulation outperforms per-voxel, since different channels encode distinct visual attributes.
Our final fusion ($\sim$8.2G) reduces the S1$\to$S2 cost by $8.9\times$ compared to the U-Net baseline ($\sim$73.4G), while achieving higher IoU and matching mIoU.

\begin{wraptable}{r}{0.45\textwidth}
\centering
\footnotesize
\caption{\textbf{Backbone \& head contributions.}}
\label{tab:abl_backbone_head}
\begin{tabular}{@{}lllcc@{}}
\toprule
Backbone & Enc. & Head (Light) & IoU & mIoU \\
\midrule
DINOv2-B & Per & w/o PA-DA & 31.76 & 20.04 \\
DINOv2-B & Per & PA-DA & 32.04 & 20.55 \\
VGGT & Uni & w/o PA-DA & 31.84 & 20.21 \\
\textbf{VGGT} & \textbf{Uni} & \textbf{PA-DA} & \textbf{33.00} & \textbf{21.08} \\
\bottomrule
\end{tabular}
\end{wraptable}

Table~\ref{tab:abl_backbone_head} disentangles whether VGGT-Occ's performance originates from the VGGT backbone or the head design.
Row 1 vs.\ 2 isolates PA-DA under a DINOv2-Base backbone: adding PA-DA brings a gain of $+0.51$ mIoU even on the weaker backbone, confirming PA-DA is backbone-agnostic.
Rows 1 and 2 vs.\ rows 3 and 4 show the contribution of upgrading from DINOv2-Base to the VGGT unified backbone ($+0.53$ mIoU).
Row 3 vs.\ 4 tests our central design hypothesis: on the same VGGT unified backbone, PA-DA exceeds standard 3D-to-2D deformable attention by $+0.87$ mIoU, confirming that injecting projection geometry into cross-attention is the key driver of accuracy.

\section{Conclusion}

We presented VGGT-Occ, a 3D occupancy prediction framework built on two principles: geometric reasoning should permeate the 2D-to-3D pipeline, and computation should be allocated by information density.
PA-DA injects projection geometry into all three stages of cross-attention: 3D offset learning, Jacobian-guided attention weighting, and view-quality gated fusion.
Coarse-guided gated fusion replaces the expensive SurroundOcc-style decoder ($\sim$73.4G) with lightweight trilinear upsampling and learned channel-wise gates ($\sim$8.2G), an $8.9\times$ reduction, by computing fusion decisions at low resolution.
With only $\sim$41M trainable head parameters, VGGT-Occ achieves state-of-the-art camera-only results on SurroundOcc-nuScenes.
PA-DA is backbone-agnostic and applicable to any deformable attention-based multi-view architecture.
Limitations include the large frozen backbone (1.2B), which increases inference memory, and the moderate input resolution ($378{\times}672$ vs.\ $900{\times}1600$ for baselines); scaling to higher resolutions would substantially increase training and inference cost. Adapting PA-DA and the density-aware coarse-guided fusion to lighter unified encoders and higher-resolution inputs are promising future directions.

\begin{ack}
\end{ack}

\newpage
\appendix

\section{Implementation and Training Details}
\label{sec:impl_details}

\subsection{Architecture Hyperparameters}

Table~\ref{tab:arch_params} lists the key architecture hyperparameters of VGGT-Occ.

\begin{table}[ht]
\centering
\caption{\textbf{Architecture hyperparameters.}}
\label{tab:arch_params}
\footnotesize
\setlength{\tabcolsep}{4pt}
\begin{tabular}{@{}lll@{}}
\toprule
Component & Parameter & Value \\
\midrule
VGGT backbone & Parameters & 1.2B (frozen) \\
& Attention blocks & 24 alternating (frame-wise + global) \\
& Input resolution & $378 {\times} 672$ \\
& Patch size & 14 \\
& Embed dim & 1024 \\
\midrule
OccHead (Coarse) & Grid size & $50{\times}50{\times}4$ (10K) \\
& Feature dim & 256 \\
& Blocks & cross, conv, cross, conv \\
\midrule
OccHead (Medium) & Grid size & $100{\times}100{\times}8$ (80K) \\
& Feature dim & 256 \\
& Blocks & cross, conv, conv \\
\midrule
OccHead (Fine) & Grid size & $200{\times}200{\times}16$ (640K) \\
& Feature dim & 64 \\
& Blocks & conv, conv \\
\midrule
PA-DA & n\_heads & 4 \\
& n\_points & 4 \\
& Offset MLP & Linear--SiLU--Linear, 3D output \\
& Jacobian bias scale $s_{h,l}$ & 0.1 \\
\midrule
Gated Fusion & Gate input & Coarse features (dim 256) \\
& Gate MLP & Linear(256,64)--SiLU--Linear(64,$C$)--Sigmoid \\
& Smoothing & Depthwise Conv3d ($k{=}3$) \\
\bottomrule
\end{tabular}
\end{table}

\subsection{Training Configuration}

Table~\ref{tab:train_params} summarizes the training configuration.

\begin{table}[ht]
\centering
\caption{\textbf{Training configuration.}}
\label{tab:train_params}
\small
\begin{tabular}{@{}ll@{}}
\toprule
Parameter & Value \\
\midrule
Dataset & SurroundOcc-nuScenes (700 train / 150 val) \\
Temporal setting & Single-frame training; multi-frame inference optional \\
Optimizer & AdamW~\cite{adamw} \\
Learning rate & $10^{-4}$ \\
Weight decay & 0.01 \\
LR schedule & Cosine annealing to $10^{-6}$ \\
Warmup & 200 iterations \\
Epochs & 20 \\
Effective batch size & 36 (3 GPU $\times$ BS3 $\times$ 4 accum) \\
Gradient clipping & 35.0 \\
Label smoothing & 0.1 \\
Mixed precision & bfloat16 \\
Distributed & DDP (3$\times$ GPU) \\
Loss & CE + sem\_scal + geo\_scal + Lov\'asz~\cite{lovasz} \\
Scale weights & [0.5, 0.75, 1.0] \\
Augmentation & Random flip, rotation $\pm5.4^\circ$, resize 0--5\% \\
EMA decay & 0.999 \\
Frozen modules & model.aggregator.* \\
\bottomrule
\end{tabular}
\end{table}

\subsection{Loss Formulation}

Let $y$ be ground-truth occupancy labels and $\hat{y}$ be predictions across three scales $s \in \{0,1,2\}$ with weights $w_s = [0.5, 0.75, 1.0]$.

\textbf{Cross-Entropy Loss.} The primary supervision signal with label smoothing $\alpha{=}0.1$:
\begin{equation}
\mathcal{L}_{\text{ce}} = \sum_s w_s \cdot \frac{1}{N_s} \sum_i \text{CE}_\alpha(\hat{y}_s^i, y_s^i)
\end{equation}

\textbf{Geometry Scene Completion Loss (geo\_scal).} Binary cross-entropy on occupied vs.\ empty, treating all semantic classes as occupied:
\begin{equation}
\mathcal{L}_{\text{geo}} = \sum_s w_s \cdot \frac{1}{N_s} \sum_i \text{BCE}(o(\hat{y}_s^i), o(y_s^i))
\end{equation}
where $o(\cdot)$ maps multi-class predictions to a binary occupancy mask.

\textbf{Semantic Scene Completion Loss (sem\_scal).} Weighted cross-entropy with inverse class-frequency weights to emphasize rare geometry classes:
\begin{equation}
\mathcal{L}_{\text{sem}} = \sum_s w_s \cdot \frac{1}{N_s} \sum_i \text{CE}_{\text{weighted}}(\hat{y}_s^i, y_s^i)
\end{equation}

\textbf{Lov\'asz-Softmax Loss.} Direct optimization of the Jaccard index:
\begin{equation}
\mathcal{L}_{\text{lovász}} = \sum_s w_s \cdot \frac{1}{C} \sum_c \overline{\Delta_{J_c}}(\hat{y}_s^c, y_s^c)
\end{equation}
where $\overline{\Delta_{J_c}}$ is the Lov\'asz extension of the Jaccard loss~\cite{lovasz} for class $c$.

\textbf{Total Loss.}
\begin{equation}
\mathcal{L} = \mathcal{L}_{\text{ce}} + \mathcal{L}_{\text{sem}} + \mathcal{L}_{\text{geo}} + \mathcal{L}_{\text{lovász}}
\end{equation}

\section{PA-DA: Mathematical Details}
\label{sec:pada_details}

We provide full mathematical derivations for the three-stage PA-DA mechanism described in Section~\ref{sec:pada} of the main text.

\subsection{Stage 1: 3D Offset Learning with Projection}

Given a 3D query $\mathbf{q} \in \mathbb{R}^d$ and its reference point $\mathbf{p}_{\text{ref}} \in \mathbb{R}^3$:

\begin{enumerate}
\item Predict independent 3D offsets (one per attention head $h$, level $l$, and sampling point $k$) via an MLP:
\begin{equation}
\boldsymbol{\Delta}^{(h,l,k)}_{3d} = \text{MLP}_{\text{offset}}(\mathbf{q})^{(h,l,k)}, \quad \boldsymbol{\Delta}^{(h,l,k)}_{3d} \in \mathbb{R}^3
\end{equation}

\item Shift the reference point by each 3D offset:
\begin{equation}
\mathbf{p}_{\text{shift}}^{(h,l,k)} = \mathbf{p}_{\text{ref}} + \boldsymbol{\Delta}^{(h,l,k)}_{3d}
\end{equation}

\item Project each shifted point to every camera $n$:
\begin{equation}
(u_n^{(h,l,k)}, v_n^{(h,l,k)}) = \pi(\mathbf{K}_n \mathbf{E}_n \mathbf{p}_{\text{shift}}^{(h,l,k)})
\end{equation}

Each 3D offset independently produces one sampling location per camera. All sampling points across heads, levels, and cameras are geometrically consistent with the same 3D reference frame.

\end{enumerate}

The projected 2D locations \emph{are} the final sampling points for each camera: we disable the default per-camera 2D sampling offsets, replacing them entirely with the geometry-consistent 3D→2D projection. This design ensures that a semantic direction (e.g., ``above the vehicle'') translates consistently across all camera views, unlike pure 2D offset learning where the same 3D direction maps to unrelated pixel shifts in each view.

\subsection{Stage 2: Jacobian attention bias}

The projection $\pi$ from 3D to each camera's image plane has an analytical Jacobian:
\begin{equation}
\mathbf{J}_n = \frac{\partial (u_n, v_n)}{\partial \mathbf{p}} \in \mathbb{R}^{2 \times 3}
\end{equation}

$\mathbf{J}_n$ captures the local sensitivity of pixel coordinates to 3D displacements. We compute its singular value decomposition:
\begin{equation}
\mathbf{J}_n = \mathbf{U} \begin{bmatrix} \sigma_{\max} & 0 & 0 \\ 0 & \sigma_{\min} & 0 \end{bmatrix} \mathbf{V}^\top
\end{equation}

The smaller non-zero singular value $\sigma_{\min}$ quantifies the weakest observable direction of 3D-to-2D information transfer (the null space, corresponding to the viewing ray, has zero sensitivity and is excluded). The Jacobian is computed in normalized image coordinates (divided by image width $W$ and height $H$), so $\sigma_{\min}$ is dimensionless. For a pinhole camera, $\sigma_{\min} = f / (Z \cdot W)$ in this normalized frame, proportional to inverse depth and giving typical values in $(0, {\sim}0.2)$ for common focal lengths and depths. For non-pinhole models (fisheye, panoramic) the SVD generalizes to more complex observation-quality patterns. When $\sigma_{\min}$ is small, the projection is degenerate: small 3D changes are imperceptible in the image, making sampled features unreliable.

For the $k$-th sampling point on camera $n$, the additive attention bias is:
\begin{equation}
b_{n,k} = s_{h,l} \cdot \log(\sigma_{\min}^{(n,k)} + \epsilon)
\end{equation}

where $s_{h,l}=0.1$ is a per-head, per-level learnable scale, and $\epsilon=10^{-5}$ prevents $\log(0)$. The modified attention weights become:
\begin{equation}
A_{n,k} = \text{softmax}\!\left(\mathbf{w}_{\text{attn}}^\top \mathbf{q} + b_{n,k}\right)
\end{equation}
where $\mathbf{w}_{\text{attn}}$ projects the query to per-point attention logits (standard deformable attention, no explicit key vectors).

For frontal cameras or nearby points with large $\sigma_{\min}$ (small depth), the bias is near zero and attention is dominated by feature similarity. For oblique or distant cameras with small $\sigma_{\min}$ (large depth), the log term is strongly negative, automatically suppressing unreliable observations.

\subsection{Stage 3: View-Quality Semantic Gate}

The full $2{\times}3$ Jacobian matrix is flattened, scaled by 1000 for numerical stability, and embedded through a 2-layer MLP with internal LayerNorm (Linear$\to$LayerNorm$\to$ReLU$\to$Linear):
\begin{equation}
\mathbf{g}_n^{\text{geo}} = \text{MLP}_{\text{geo}}\!\big(\mathrm{flatten}(\mathbf{J}_n) \times 1000\big) \in \mathbb{R}^{C}
\end{equation}

This geometric encoding is concatenated with the query feature $\mathbf{f}$ and fused through a second 2-layer MLP (Linear$\to$LayerNorm$\to$ReLU$\to$Linear$\to$Sigmoid) to produce per-channel fusion weights:
\begin{equation}
\boldsymbol{\gamma}_n = \text{MLP}_{\text{gate}}\!([\,\mathbf{f} \;\|\; \mathbf{g}_n^{\text{geo}}\,]) \in \mathbb{R}^{C}
\end{equation}
where $\|$ denotes channel-wise concatenation and the final activation is sigmoid.

Cross-camera features are fused via per-channel, per-camera gating, normalized by the sum of gates:
\begin{equation}
\hat{\mathbf{f}} = \frac{\sum_n \boldsymbol{\gamma}_n \odot \mathbf{v}_n}{\sum_n \boldsymbol{\gamma}_n},
\end{equation}
where $\mathbf{v}_n$ is the sampled feature from camera $n$. The denominator $\sum_n \boldsymbol{\gamma}_n$ normalizes the fusion, so that the output magnitude is independent of the number of cameras.

\section{Gate Heatmap Analysis}
\begin{figure}[htbp]
  \centering
  \includegraphics[width=\linewidth]{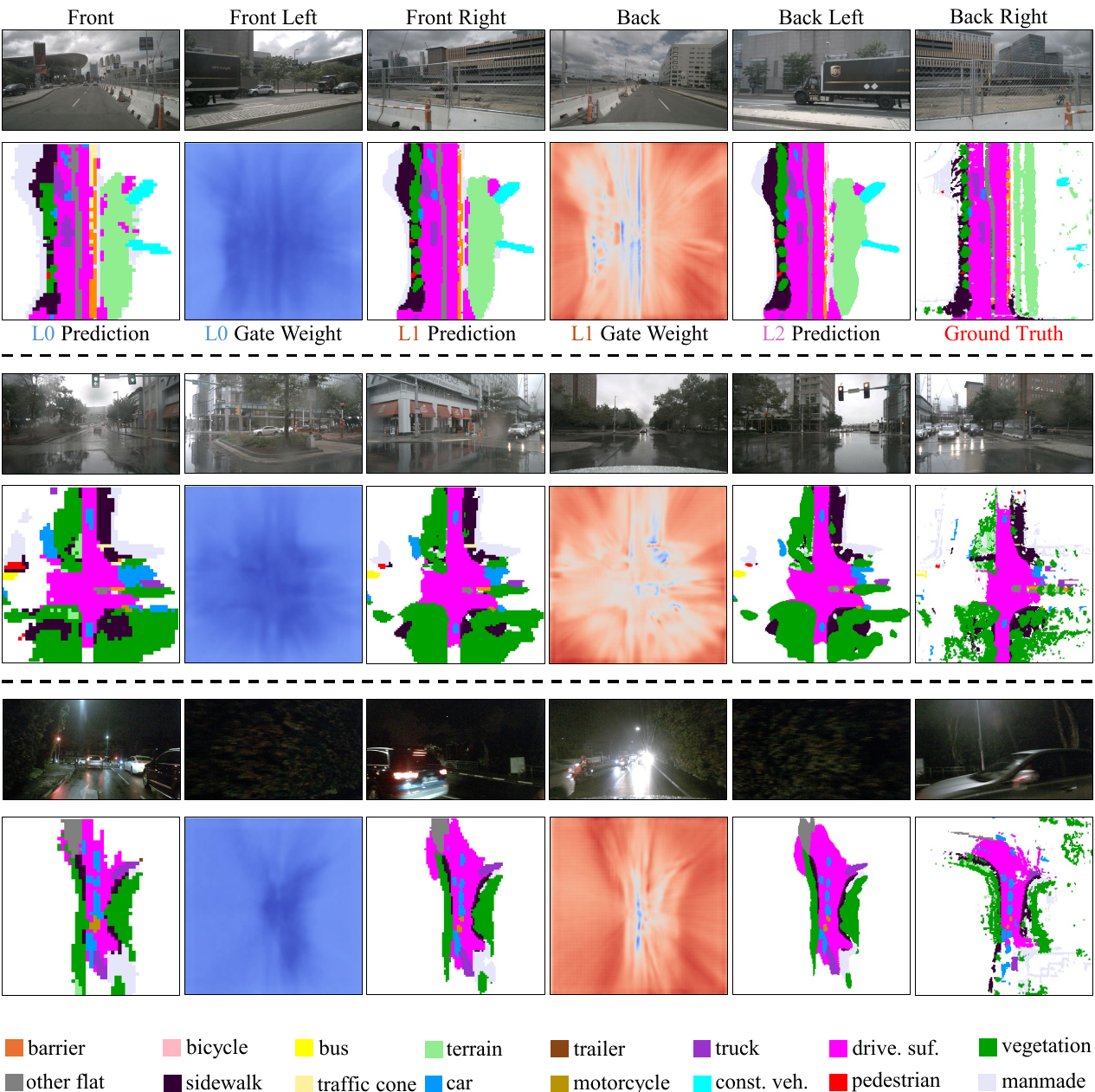}
  \caption{\textbf{Gate heatmap under challenging conditions.} Daytime clutter (top), heavy rain (middle), and nighttime (bottom). Warmer colors indicate higher reliance on coarse-level semantic information, while cooler colors represent a shift toward fine-scale structural details. The gating mechanism dynamically adapts to both environmental noise and local geometric complexity.}
  \label{fig:gate_supp}
\end{figure}

Figure~\ref{fig:gate_supp} visualizes the coarse-guided gating under three challenging conditions: daytime clutter (construction fences), heavy rain, and nighttime.
The $L_0$ gate consistently trusts fine-scale features (cooler blue tones), while the $L_1$ gate relies more on coarse semantics (warmer tones) to maintain structural coherence.
Under adverse conditions (rain, night), the $L_1$ gate shifts toward even stronger coarse reliance in feature-poor regions to compensate for degraded local textures.
Conversely, in the cluttered daytime scene, cooler streaks appear at fence and barrier boundaries, indicating the network defers to fine-scale features to resolve thin geometric structures.
This adaptive behavior confirms that coarse-guided gating balances semantic stability with spatial precision across diverse environments.

\section{Additional Qualitative Results}

\begin{figure}[htbp]
  \centering
  \includegraphics[width=\linewidth]{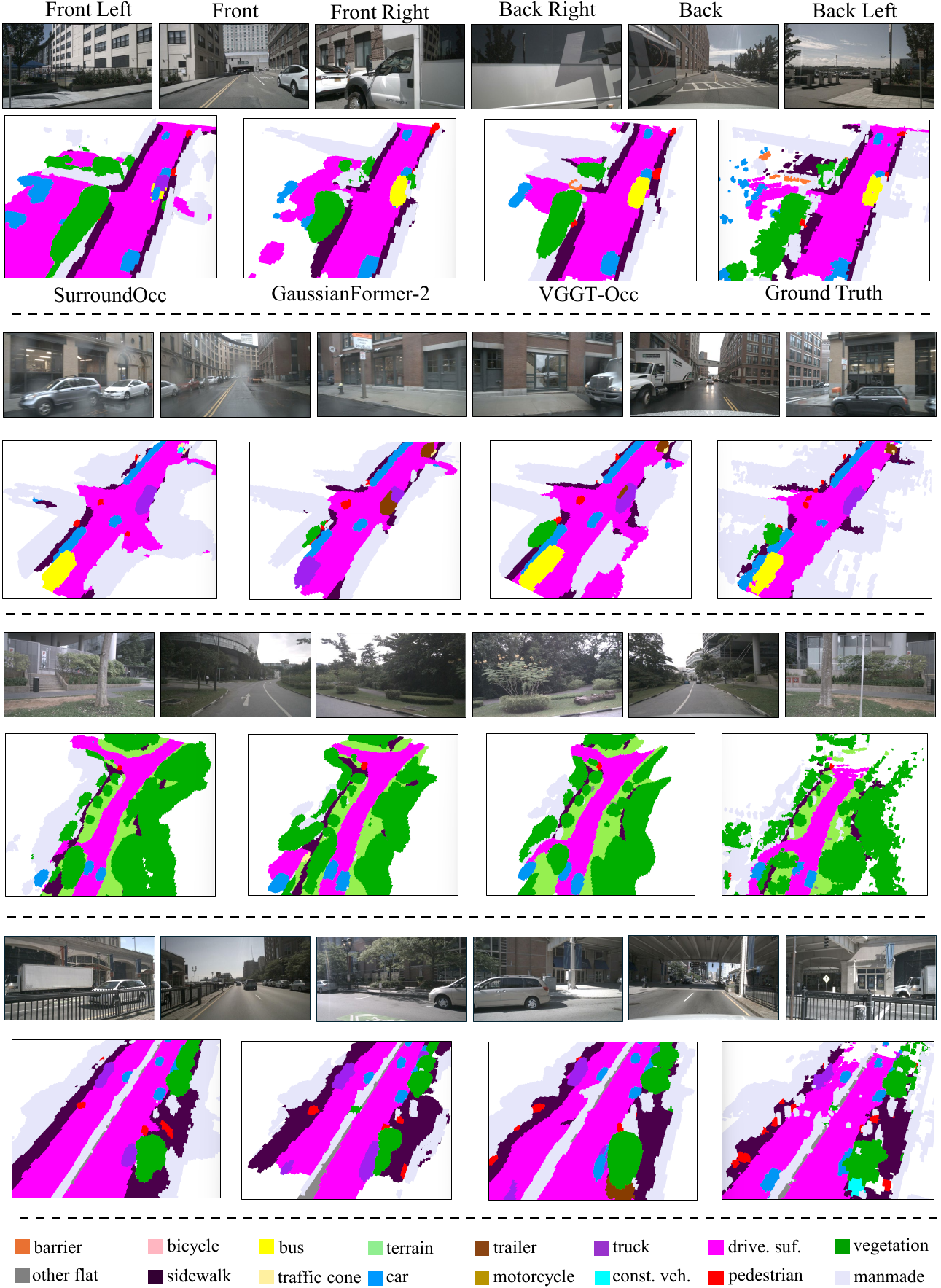}
  \caption{\textbf{Qualitative comparison on SurroundOcc-nuScenes.}
  Qualitative results of VGGT-Occ compared with state-of-the-art methods. VGGT-Occ produces finer geometric structures and more accurate semantic boundaries, aligning much more closely with the ground truth in complex scenarios.}
  \label{fig:qual_supp}
\end{figure}

Figure~\ref{fig:qual_supp} provides a qualitative comparison on challenging SurroundOcc-nuScenes scenes.
With end-to-end geometry injection (cross-view unified encoding via VGGT, 3D offset projection, Jacobian-guided attention, and view-quality gated fusion), VGGT-Occ produces more precise geometric structures and cleaner semantic boundaries, particularly in cluttered or distant regions where purely data-driven methods tend to blur or fragment.
This aligns with our quantitative findings: geometry-grounded attention preserves fine-grained sampling in high-quality views while suppressing unreliable observations.

\bibliographystyle{plainnat}
\bibliography{references}

@inproceedings{surroundocc,
  title={{SurroundOcc}: Multi-camera {3D} Occupancy Prediction for Autonomous Driving},
  author={Wei, Yi and Zhao, Linqing and Zheng, Wenzhao and Zhu, Zheng and Zhou, Jie and Lu, Jiwen},
  booktitle={ICCV},
  year={2023}
}

@inproceedings{occ3d,
  title={{Occ3D}: A Large-Scale {3D} Occupancy Prediction Benchmark for Autonomous Driving},
  author={Tian, Xiaoyu and Jiang, Tao and Yun, Longfei and Wang, Yue and Wang, Yilun and Zhao, Hang},
  booktitle={NeurIPS},
  year={2023}
}

@inproceedings{occformer,
  title={{OccFormer}: Dual-path Transformer for Vision-based {3D} Semantic Occupancy Prediction},
  author={Zhang, Yunpeng and Zhu, Zheng and Du, Dalong},
  booktitle={ICCV},
  year={2023}
}

@inproceedings{gaussianformer2,
  title={{GaussianFormer-2}: Probabilistic {G}aussian Superposition for Efficient {3D} Occupancy Prediction},
  author={Huang, Yuanhui and Thammatadatrakoon, Amonnut and Zheng, Wenzhao and Zhang, Yunpeng and Du, Dalong and Lu, Jiwen},
  booktitle={CVPR},
  year={2025}
}

@inproceedings{quadricformer,
  title={{QuadricFormer}: Scene as Superquadrics for {3D} Semantic Occupancy Prediction},
  author={Zuo, Sicheng and Zheng, Wenzhao and Han, Xiaoyong and Yang, Longchao and Pan, Yong and Lu, Jiwen},
  booktitle={NeurIPS},
  year={2025}
}

@inproceedings{bevformer,
  title={{BEVFormer}: Learning Bird's-Eye-View Representation from Multi-Camera Images via Spatiotemporal Transformers},
  author={Li, Zhiqi and Wang, Wenhai and Li, Hongyang and Xie, Enze and Sima, Chonghao and Lu, Tong and Qiao, Yu and Dai, Jifeng},
  booktitle={ECCV},
  year={2022}
}

@article{sparse4d,
  title={{Sparse4D}: Multi-view {3D} Object Detection with Sparse Spatial-Temporal Fusion},
  author={Lin, Xuewu and Lin, Tianwei and Pei, Zixiang and Huang, Lichao and Su, Zhizhong},
  journal={arXiv preprint arXiv:2211.10581},
  year={2022}
}

@inproceedings{vggt,
  title={{VGGT}: Visual Geometry Grounded Transformer},
  author={Wang, Jianyuan and Chen, Minghao and Karaev, Nikita and Vedaldi, Andrea and Rupprecht, Christian and Novotny, David},
  booktitle={CVPR},
  year={2025}
}

@inproceedings{tpvformer,
  title={Tri-Perspective View for Vision-Based {3D} Semantic Occupancy Prediction},
  author={Huang, Yuanhui and Zheng, Wenzhao and Zhang, Yunpeng and Zhou, Jie and Lu, Jiwen},
  booktitle={CVPR},
  year={2023}
}

@inproceedings{gaussrender,
  title={{GaussRender}: Learning {3D} Occupancy with {G}aussian Rendering},
  author={Chambon, Lo{\"i}ck and Zablocki, Eloi and Boulch, Alexandre and Chen, Micka{\"e}l and Cord, Matthieu},
  booktitle={ICCV},
  year={2025}
}

@inproceedings{deformabledetr,
  title={Deformable {DETR}: Deformable Transformers for End-to-End Object Detection},
  author={Zhu, Xizhou and Su, Weijie and Lu, Lewei and Li, Bin and Wang, Xiaogang and Dai, Jifeng},
  booktitle={ICLR},
  year={2021}
}

@inproceedings{petr,
  title={{PETR}: Position Embedding Transformation for Multi-View {3D} Object Detection},
  author={Liu, Yingfei and Wang, Tiancai and Zhang, Xiangyu and Sun, Jian},
  booktitle={ECCV},
  year={2022}
}

@inproceedings{lovasz,
  title={The {L}ov{\'a}sz-Softmax Loss: A Tractable Surrogate for the Optimization of the Intersection-Over-Union Measure in Neural Networks},
  author={Berman, Maxim and Triki, Amal Rannen and Blaschko, Matthew B.},
  booktitle={CVPR},
  year={2018}
}

@article{bevdet,
  title={{BEVDet}: High-Performance Multi-Camera {3D} Object Detection in Bird-Eye-View},
  author={Huang, Junjie and Huang, Guan and Zhu, Zheng and Ye, Yun and Du, Dalong},
  journal={arXiv preprint arXiv:2112.11790},
  year={2021}
}

@inproceedings{nuscenes,
  title={{nuScenes}: A Multimodal Dataset for Autonomous Driving},
  author={Caesar, Holger and Bankiti, Varun and Lang, Alex H. and Vora, Sourabh and Liong, Venice Erin and Xu, Qiang and Krishnan, Anush and Pan, Yu and Baldan, Giancarlo and Beijbom, Oscar},
  booktitle={CVPR},
  year={2020}
}

@article{dinov2,
  title={{DINOv2}: Learning Robust Visual Features without Supervision},
  author={Oquab, Maxime and Darcet, Timoth{\'e}e and Moutakanni, Th{\'e}o and Vo, Huy and Szafraniec, Marc and Khalidov, Vasil and Fernandez, Pierre and Haziza, Daniel and Massa, Francisco and El-Nouby, Alaaeldin and others},
  journal={TMLR},
  year={2024}
}

@inproceedings{gaussianformer,
  title={{GaussianFormer}: Scene as {G}aussians for Vision-Based {3D} Semantic Occupancy Prediction},
  author={Huang, Yuanhui and Zheng, Wenzhao and Zhang, Yunpeng and Zhou, Jie and Lu, Jiwen},
  booktitle={ECCV},
  year={2024}
}

@inproceedings{adamw,
  title={Decoupled Weight Decay Regularization},
  author={Loshchilov, Ilya and Hutter, Frank},
  booktitle={ICLR},
  year={2019}
}

@inproceedings{vit,
  title={An Image is Worth 16x16 Words: Transformers for Image Recognition at Scale},
  author={Dosovitskiy, Alexey and Beyer, Lucas and Kolesnikov, Alexander and Weissenborn, Dirk and Zhai, Xiaohua and Unterthiner, Thomas and Dehghani, Mostafa and Minderer, Matthias and Heigold, Georg and Gelly, Sylvain and others},
  booktitle={ICLR},
  year={2021}
}

@inproceedings{monoscene,
  title={{MonoScene}: Monocular {3D} Semantic Scene Completion},
  author={Cao, Anh-Quan and de Charette, Raoul},
  booktitle={CVPR},
  year={2022}
}

@article{flashocc,
  title={{FlashOcc}: Fast and Memory-Efficient Occupancy Prediction via Channel-to-Height Plugin},
  author={Yu, Zichen and Shu, Changyong and Deng, Jiajun and Lu, Kangjie and Liu, Zongdai and Yu, Jiangyong and Yang, Dawei and Li, Hui and Chen, Yan},
  journal={arXiv preprint arXiv:2311.12058},
  year={2023}
}

@inproceedings{ctfocc,
  title={{CTF-Occ}: Coarse-to-Fine {3D} Occupancy Prediction},
  author={Tian, Xiaoyu and Jiang, Tao and Yun, Longfei and Mao, Yucheng and Yang, Huitong and Wang, Yue and Wang, Yilun and Zhao, Hang},
  booktitle={NeurIPS},
  year={2023}
}

@inproceedings{convnext,
  title={A {ConvNet} for the 2020s},
  author={Liu, Zhuang and Mao, Hanzi and Wu, Chao-Yuan and Feichtenhofer, Christoph and Darrell, Trevor and Xie, Saining},
  booktitle={CVPR},
  year={2022}
}

@article{deng2025best3dscenerepresentation,
  title={What Is The Best {3D} Scene Representation for Robotics? From Geometric to Foundation Models},
  author={Deng, Tianchen and Pan, Yue and Yuan, Shenghai and Li, Dong and Wang, Chen and Li, Mingrui and Chen, Long and Xie, Lihua and Wang, Danwei and Wang, Jingchuan and Civera, Javier and Wang, Hesheng and Chen, Weidong},
  journal={arXiv preprint arXiv:2512.03422},
  year={2025}
}

@inproceedings{pan2024renderocc,
  title={{RenderOcc}: Vision-Centric {3D} Occupancy Prediction with {2D} Rendering Supervision},
  author={Pan, Mingjie and Liu, Jiaming and Zhang, Renrui and Huang, Peixiang and Li, Xiaoqi and Xie, Hongwei and Wang, Bing and Liu, Li and Zhang, Shanghang},
  booktitle={ICRA},
  year={2024}
}

@inproceedings{li2023fbocc,
  title={{FB-OCC}: {3D} Occupancy Prediction based on Forward-Backward View Transformation},
  author={Li, Zhiqi and Yu, Zhiding and Austin, David and Fang, Mingsheng and Lan, Shiyi and Kautz, Jan and Alvarez, Jose M},
  booktitle={CVPR Workshop on End-to-End Autonomous Driving},
  year={2023}
}

@inproceedings{wang2024panoocc,
  title={{PanoOcc}: Unified Occupancy Representation for Camera-based {3D} Panoptic Segmentation},
  author={Wang, Yuqi and Chen, Yuntao and Liao, Xingyu and Fan, Lue and Zhang, Zhaoxiang},
  booktitle={CVPR},
  year={2024}
}

@inproceedings{liu2024sparseocc,
  title={Fully Sparse {3D} Occupancy Prediction},
  author={Liu, Haisong and Chen, Yang and Wang, Haiguang and Yang, Zetong and Li, Tianyu and Zeng, Jia and Chen, Li and Li, Hongyang and Wang, Limin},
  booktitle={ECCV},
  year={2024}
}

@inproceedings{zhang2025sqs,
  title={{SQS}: Enhancing Sparse Perception Models via Query-based Splatting in Autonomous Driving},
  author={Zhang, Haiming and Zhu, Yiyao and Zhou, Wending and Yan, Xu and Cai, Yingjie and Liu, Bingbing and Cui, Shuguang and Li, Zhen},
  booktitle={NeurIPS},
  year={2025}
}

@inproceedings{jiang2024far3d,
  title={{Far3D}: Expanding the Horizon for Surround-view {3D} Object Detection},
  author={Jiang, Xiaohui and Li, Shuailin and Liu, Yingfei and Wang, Shihao and Jia, Fan and Wang, Tiancai and Han, Lijin and Zhang, Xiangyu},
  booktitle={AAAI},
  year={2024}
}

@inproceedings{oh2025protoocc,
  title={{3D} Occupancy Prediction with Low-Resolution Queries via Prototype-aware View Transformation},
  author={Oh, Gyeongrok and Kim, Sungjune and Ko, Heeju and Chi, Hyung-gun and Kim, Jinkyu and Lee, Dongwook and Ji, Daehyun and Choi, Sungjoon and Jang, Sujin and Kim, Sangpil},
  booktitle={CVPR},
  year={2025}
}

@inproceedings{zhu2026drocc,
  title={{Dr.Occ}: Depth- and Region-Guided {3D} Occupancy from Surround-View Cameras for Autonomous Driving},
  author={Zhu, Xubo and Zhang, Haoyang and He, Fei and Wu, Rui and Shan, Yanhu and Yang, Wen and Yu, Huai},
  booktitle={CVPR},
  year={2026}
}

@inproceedings{shi2026bepo,
  title={{BePo}: Dual Representation for {3D} Occupancy Prediction},
  author={Shi, Yunxiao and Cai, Hong and Jeong, Jisoo and Zhu, Yinhao and Han, Shizhong and Ansari, Amin and Porikli, Fatih},
  booktitle={CVPR Workshop on Autonomous Driving},
  year={2026}
}

@inproceedings{zhao2026gaussianformer3d,
  title={{GaussianFormer3D}: Multi-Modal {G}aussian-based Semantic Occupancy Prediction with {3D} Deformable Attention},
  author={Zhao, Lingjun and Wei, Sizhe and Hays, James and Gan, Lu},
  booktitle={ICRA},
  year={2026}
}

@inproceedings{huang2024selfocc,
  title={{SelfOcc}: Self-Supervised Vision-Based {3D} Occupancy Prediction},
  author={Huang, Yuanhui and Zheng, Wenzhao and Zhang, Borui and Zhou, Jie and Lu, Jiwen},
  booktitle={CVPR},
  year={2024}
}

@inproceedings{zuo2025gaussianworld,
  title={{GaussianWorld}: {G}aussian World Model for Streaming {3D} Occupancy Prediction},
  author={Zuo, Sicheng and Zheng, Wenzhao and Huang, Yuanhui and Zhou, Jie and Lu, Jiwen},
  booktitle={CVPR},
  year={2025}
}

@article{zhang2026survey,
  title={Vision-based {3D} occupancy prediction in autonomous driving: a review and outlook},
  author={Zhang, Yanan and Zhang, Jinqing and Wang, Zengran and Xu, Junhao and Huang, Di},
  journal={Frontiers of Computer Science},
  volume={20},
  pages={2001301},
  year={2026}
}

@article{xu2025survey,
  title={A survey on occupancy perception for autonomous driving: The information fusion perspective},
  author={Xu, Huaiyuan and Chen, Junliang and Meng, Shiyu and Wang, Yi and Chau, Lap-Pui},
  journal={Information Fusion},
  volume={114},
  pages={102671},
  year={2025}
}

@inproceedings{detr3d,
  title={{DETR3D}: {3D} Object Detection from Multi-view Images via {3D}-to-{2D} Queries},
  author={Wang, Yue and Guizilini, Vitor Campagnolo and Zhang, Tianyuan and Wang, Yilun and Zhao, Hang and Solomon, Justin},
  booktitle={CoRL},
  year={2021}
}

@inproceedings{lss,
  title={Lift, Splat, Shoot: Encoding Images From Arbitrary Camera Rigs by Implicitly Unprojecting to {3D}},
  author={Philion, Jonah and Fidler, Sanja},
  booktitle={ECCV},
  year={2020}
}

@article{3dgs,
  title={{3D} {G}aussian Splatting for Real-Time Radiance Field Rendering},
  author={Kerbl, Bernhard and Kopanas, Georgios and Leimk{\"u}hler, Thomas and Drettakis, George},
  journal={ACM Trans. Graph.},
  volume={42},
  number={4},
  year={2023}
}

@article{sscbench,
  title={{SSCBench}: A Large-Scale {3D} Semantic Scene Completion Benchmark for Autonomous Driving},
  author={Li, Yiming and Li, Sihang and Liu, Xinhao and Gong, Moonjun and Li, Kenan and Chen, Nuo and Wang, Zijun and Li, Zhiheng and Jiang, Tao and Yu, Fisher and Wang, Yue and Zhao, Hang and Yu, Zhiding and Feng, Chen},
  journal={arXiv preprint arXiv:2306.09001},
  year={2023}
}

@inproceedings{depthanything,
  title={Depth Anything: Unleashing the Power of Large-Scale Unlabeled Data},
  author={Yang, Lihe and Kang, Bingyi and Huang, Zilong and Xu, Xiaogang and Feng, Jiashi and Zhao, Hengshuang},
  booktitle={CVPR},
  year={2024}
}

@inproceedings{oft,
  title={Orthographic Feature Transform for Monocular {3D} Object Detection},
  author={Roddick, Thomas and Kendall, Alex and Cipolla, Roberto},
  booktitle={BMVC},
  year={2019}
}

@article{unipr-3d,
  title={{UniPR-3D}: Towards Universal Visual Place Recognition with Visual Geometry Grounded Transformer},
  author={Deng, Tianchen and Chen, Xun and Li, Ziming and Shen, Hongming and Wang, Danwei and Civera, Javier and Wang, Hesheng},
  journal={arXiv preprint arXiv:2512.21078},
  year={2025}
}

@article{relocvggt,
  title={{Reloc-VGGT}: Visual Re-localization with Geometry Grounded Transformer},
  author={Deng, Tianchen and Wu, Wenhua and Wu, Kunzhen and Wang, Guangming and Zhu, Siting and Yuan, Shenghai and Chen, Xun and Shen, Guole and Liu, Zhe and Wang, Hesheng},
  journal={arXiv preprint arXiv:2512.21883},
  year={2025}
}

@article{deng2024compact,
  title={Compact {3D} {G}aussian Splatting for Dense Visual {SLAM}},
  author={Deng, Tianchen and Chen, Yaohui and Zhang, Leyan and Yang, Jianfei and Yuan, Shenghai and Liu, Jiuming and Wang, Danwei and Wang, Hesheng and Chen, Weidong},
  journal={arXiv preprint arXiv:2403.11247},
  year={2024}
}

@inproceedings{qian2026splatssc,
  title={{SplatSSC}: Decoupled Depth-Guided {G}aussian Splatting for Semantic Scene Completion},
  author={Qian, Rui and Cao, Haozhi and Deng, Tianchen and Yuan, Shenghai and Xie, Lihua},
  booktitle={Proceedings of the AAAI Conference on Artificial Intelligence},
  volume={40},
  number={10},
  pages={8520--8528},
  year={2026}
}

@article{qian2025tgsformer,
  title={{TGSFormer}: Scalable Temporal {G}aussian Splatting for Embodied Semantic Scene Completion},
  author={Qian, Rui and Cao, Haozhi and Deng, Tianchen and Hu, Tianxin and Guo, Weixiang and Yuan, Shenghai and Xie, Lihua},
  journal={arXiv preprint arXiv:2512.00300},
  year={2025}
}

@inproceedings{ma2024cam4docc,
  title={{Cam4DOcc}: Benchmark for Camera-Only {4D} Occupancy Forecasting in Autonomous Driving Applications},
  author={Ma, Junyi and Chen, Xieyuanli and Huang, Jiawei and Xu, Jingyi and Luo, Zhen and Xu, Jintao and Gu, Weihao and Ai, Rui and Wang, Hesheng},
  booktitle={Proceedings of the IEEE/CVF Conference on Computer Vision and Pattern Recognition},
  pages={21486--21495},
  year={2024}
}

\end{document}